\documentclass[preprint]{article}
\usepackage{amsmath,amssymb,graphicx,xcolor}
\usepackage{neurips_2024}
\usepackage[utf8]{inputenc} 
\usepackage[T1]{fontenc}    
\usepackage{hyperref}       
\usepackage{url}            
\usepackage{booktabs}       
\usepackage{amsfonts}       
\usepackage{nicefrac}       
\usepackage{microtype}      
\usepackage{wrapfig}
\usepackage{tabularx}
\usepackage{multirow}

\usepackage{tikz}

\usepackage{graphicx}
\usepackage{eso-pic} 

\AddToShipoutPictureBG*{%
  \put(50,640){\includegraphics[width=2.0cm]{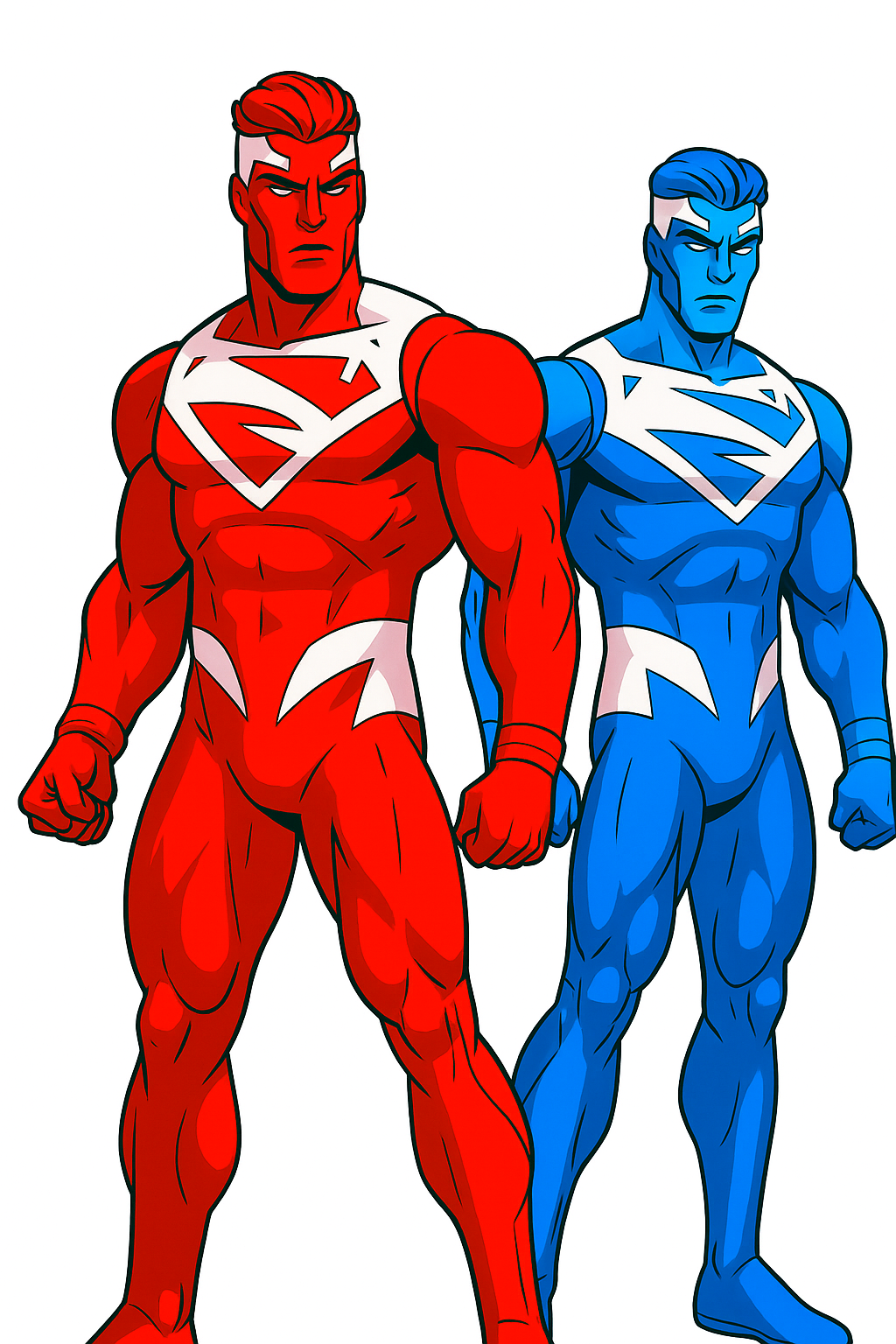}} 
}

\title{%
  \textit{LegiGPT}: Party Politics in Transport Policy with a Large Language Model 
}

\author{
Hyunsoo Yun$^{1}$, Eun Hak Lee$^{2}$\thanks{Corresponding author.} \\
$^{1}$Seoul National University \\
$^{2}$Texas A\&M Transportation Institute \\
{\tt\small yoonhs1113@snu.ac.kr, e-lee@tti.tamu.edu}
}

\begin{document}
\maketitle

\begin{abstract}
Given the significant influence of lawmakers’ political ideologies on legislative decision-making, understanding their impact on policymaking is critically important. We introduce a novel framework, \textbf{LegiGPT}, which integrates a large language model (LLM) with explainable artificial intelligence (XAI) to analyze transportation-related legislative proposals. \textbf{LegiGPT} employs a multi-stage filtering and classification pipeline using zero-shot prompting with GPT-4. Using legislative data from South Korea’s 21st National Assembly, we identify key factors—including sponsor characteristics, political affiliations, and geographic variables—that significantly influence transportation policymaking. The LLM was used to classify transportation-related bill proposals through a stepwise filtering process based on keywords, phrases, and contextual relevance. XAI techniques were then applied to examine relationships between party affiliation and associated attributes. The results reveal that the number and proportion of conservative and progressive sponsors, along with district size and electoral population, are critical determinants shaping legislative outcomes. These findings suggest that both parties contributed to bipartisan legislation through different forms of engagement, such as initiating or supporting proposals. This integrated approach provides a valuable tool for understanding legislative dynamics and guiding future policy development, with broader implications for infrastructure planning and governance.

\end{abstract}


\newpage
\section{Introduction}
Transportation legislation significantly shapes infrastructure, economic growth, and public welfare. In democratic systems, political ideologies influence these policies by guiding funding decisions and strategic priorities—e.g., rural vs. urban investments or highways vs. public transit~\cite{sorensen2014political}. In South Korea, this ideological divide is spatially evident: conservatives dominate in Gyeongsang, progressives in Jeolla~\cite{chae2008conservatives, yun2024}. Within Seoul, Gangnam supports conservatives, while other districts lean progressive. These divides manifest not only in policy content but also in legislative language, reflecting party values and electoral goals.

Recent studies highlight how political dynamics shape transportation legislation, particularly in funding priorities and project approvals~\cite{klein2022political, legacy2019consensus, mcandrews2015politics, sciara2024state}. For example, federal budget analyses show that states aligned with the ruling party often receive more infrastructure funding. A 2024 report found that 73\% of funds (\textasciitilde\$7.8B) went to Democrat-led states, despite their being fewer in number~\cite{christenson2024}. Ideological divides also affect support for sustainable initiatives: conservative administrations tend to boost rural infrastructure, while progressive ones favor electric vehicles and urban transit~\cite{usafacts2024}. Voting patterns further illustrate this divide. The 2021 Infrastructure Investment and Jobs Act, which allocated \$1.2 trillion for projects like high-speed rail, was widely supported by progressives but opposed by conservatives prioritizing highway expansion and fossil fuel infrastructure~\cite{usdot2023}.

As such, statistical trends show that political ideologies shape both the content and direction of transportation legislation, leading to disparities in resource allocation and reflecting deeper ideological divisions. Prior studies have examined this relationship through surveys. For instance, Klein et al.~\cite{klein2022political} found that conservatives tend to support highway expansion, while progressives favor public transit. Christiansen~\cite{christiansen2018public} further showed that opposition to restrictive policies like local tolls can erode satisfaction with democratic institutions, highlighting the link between policy design, public attitudes, and political legitimacy. While informative, these survey-based approaches face challenges, including low response rates, bias, and resource constraints.

Given these challenges, many studies have used text mining to extract insights from large-scale unstructured data in the transportation sector. Applications range from analyzing passenger reviews to inform resilient policies~\cite{park2022toward}, to identifying key planning factors from policy documents~\cite{chowdhury2023investigation}, and measuring user satisfaction via social media sentiment~\cite{zou2024b}. Chen et al.~\cite{chen2024exploring} also traced shifts in port policy using text mining. However, traditional methods often rely on simplified linguistic features, limiting their ability to capture complex structures and evolving terminology.

Large language models (LLMs) provide an alternative. Their pre-trained understanding enables zero-shot analysis of legislative texts without labeled data~\cite{kuila2024deciphering, leong2024metroberta, lim2025large, tornberg2024large, kim2025linear}. LLMs offer three main advantages: (1) efficient processing of large unstructured datasets, (2) reduced human bias, and (3) the ability to detect latent patterns and contextual nuances~\cite{wang2025, wei2022b}. They also generalize across tasks without task-specific training~\cite{wei2022a, zhang2024}. Despite these benefits, the use of LLMs to systematically analyze how political ideologies influence transportation legislation remains limited—highlighting the need for further research in this area.

This study aims to understand how political ideology shapes the proposal patterns, sponsor composition, and approval outcomes of transportation-related legislation. To achieve this, we developed \textbf{LegiGPT} (Legislative GPT), which integrates two key components: (1) a LLM is used to extract and refine transportation-related bills from legislative texts through a multi-stage filtering process; and (2) explainable AI (XAI) techniques are applied to model and interpret the relationships between political party affiliation and key legislative, demographic, and geographic attributes. The results reveal the most influential attributes and offer insights into their interactions. This integrated approach demonstrates the potential of AI-driven methodologies for uncovering latent patterns in transportation policymaking influenced by political ideologies.

\begin{itemize}
  \item We develop \textbf{LegiGPT}, a novel framework that integrates LLMs with explainable artificial intelligence (XAI) techniques to analyze transportation-related legislative proposals.
  
  \item We propose a multi-stage classification approach that leverages zero-shot prompting to filter transportation bills from raw legislative texts without requiring labeled data.
  
  \item We reveal key political and geographic factors, such as party affiliation, sponsor ideology, and electoral population, that shape legislative outcomes in transportation policymaking.
\end{itemize}

\section{Data}

The open data portal in South Korea provides a centralized platform for accessing a wide range of government datasets~\cite{data_go_kr}. It supports transparency and innovation by offering open data across various sectors, including legislation, transportation, and public administration.

Among the available datasets, two were used in this study. First, the National Assembly's legislative activity information was utilized. The legislative process in the Korean National Assembly comprises four main steps: (1) bill introduction, (2) review by the relevant standing committee, (3) review by the Legislation and Judiciary Committee (LJC), and (4) floor debate and voting.

Initially, a bill is introduced by a legislator, referred to as the sponsor. Other lawmakers who support the bill are designated as co-sponsors. The proposed bill is then reviewed by the standing committee specializing in the subject to assess its necessity and relevance. Bills approved by the committee are passed on to the LJC, which examines their legal structure and terminology. If approved, the bill proceeds to floor debate and is voted on. A bill is enacted into law if it receives a majority vote and is signed by the president. At each stage, a bill may advance or be dropped. The legislative dataset includes attributes, i.e., bill ID, bill title, description, sponsors, and review outcomes for each stage.

Second, election data—comprising voting and ballot-counting results—was used. This dataset provides granular insights into electoral outcomes by district, including voter turnout and party-level vote shares. It also includes candidate names, the number of electors, voters, votes received, votes not received, and abstentions, aggregated at district, regional, and national levels.

\begin{wrapfigure}{r}{0.48\textwidth}
  \centering
  \vspace{0pt}
  \includegraphics[width=0.46\textwidth]{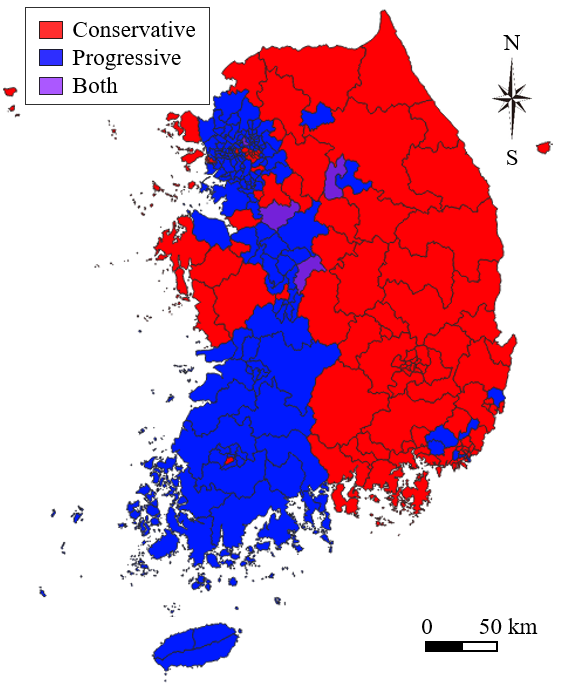}
  \vspace{-6pt}
  \caption{\small Political parties by district.}
  \vspace{-20pt}
  \label{fig:party_district_map}
\end{wrapfigure}

Both datasets focus on the legislative activities of South Korea’s 21st National Assembly, covering the period from June 2020 to May 2024. This timeframe encompasses all bills proposed, reviewed, and processed during the 21st Assembly, offering a comprehensive view of legislative dynamics over the four-year term.

Over this period, 23,655 bills were proposed, of which 1,359 were approved. The approval process includes committee review, LJC assessment, and a final plenary vote. Only bills that pass the final plenary session are enacted. There are 253 electoral districts in South Korea. The National Assembly is composed of 322 members—266 elected from districts and 56 proportional representatives. Since vacancies may occur mid-term, by-elections are held, and new lawmakers are elected, which results in the number of district legislators exceeding the number of electoral districts. Each bill may involve multiple lawmakers, leading to a total of 297,155 lawmaker-bill participation samples, calculated as the product of proposals and their sponsors. The distribution of political parties by electoral district is visualized in Figure~\ref{fig:party_district_map}.

\section{Methods}

\subsection{Large Language Model}

We developed \textbf{LegiGPT}, a legislative analysis pipeline powered by GPT-4, to process and refine bill data related to transportation. Here, “transportation-related” refers to legislative content whose primary objective involves the planning, regulation, operation, or development of transportation systems. This includes policies or services related to road infrastructure, traffic management, public transit (e.g., buses, subways), mobility services, vehicle operations, and associated facilities such as signals, bike lanes, and transportation hubs. Bills addressing adjacent areas—such as logistics, pedestrian infrastructure, or transportation safety—were also considered transportation-related if the legislative intent was clearly linked to the movement of people or goods.

\textbf{LegiGPT} consisted of four stages: (1) keyword extraction, (2) keyword-based bill selection, (3) sentence-based bill selection, and (4) context-based bill selection. GPT-4 was used to perform filtering and translation tasks with the following hyperparameters: \texttt{temperature = 0.2}, \texttt{max\_tokens = 256}, \texttt{top\_p = 1.0}, with default values for all other parameters. At each stage, a randomly selected 10\% sample was manually reviewed to ensure accuracy and consistency.

\textbf{\underline{First}}, keywords were extracted through word frequency analysis. The LLM was instructed to count all nouns in the description field of the bill data. Since the original data was in Korean, the extracted keywords were translated into English. This translation helps mitigate misinterpretation caused by linguistic differences, enabling more accurate classification. However, this translation step may not be necessary in all cases; its utility depends on the linguistic distance between source and target languages~\cite{diandaru2024could}. \textbf{\underline{Second}}, bill data containing at least one of the extracted keywords was filtered. The keywords and bill data were uploaded to the LLM, and prompts were used to extract all records where the bill summary field included one or more keywords. This step captures a broad set of potentially relevant bills, including those where transportation terms appear incidentally or metaphorically. \textbf{\underline{Third}}, sentence-based bill selection was applied to refine the results. The LLM evaluated individual sentences to determine whether the context of keywords genuinely related to transportation. For instance, a keyword may appear in a sentence, but the overall focus might pertain to a different domain. Only bill records with at least one sentence directly relevant to transportation policy were retained. \textbf{\underline{Finally}}, context-based bill selection was performed to ensure legislative intent truly centered on transportation. The LLM assessed the broader context surrounding each bill, evaluating whether its primary objective aligned with transportation-related goals. Even when a sentence mentioned transportation, bills were excluded if their legislative focus was unrelated to transport systems or infrastructure.

\begin{figure}[htbp]
  \centering
  \includegraphics[width=1.0\linewidth]{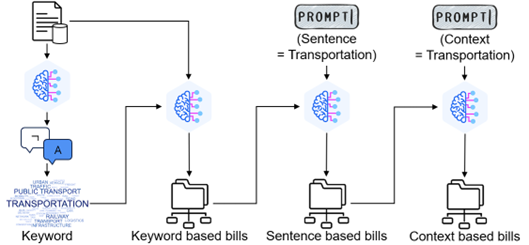}
  \caption{Large Language Model Framework for Bill data analysis.}
  \label{fig:llm_pipeline}
\end{figure}

This four-step method begins with broad keyword selection and progressively applies contextual refinement. The structured filtering pipeline balances efficiency with precision, ensuring that selected texts reflect substantive transportation policy content. The overall process is illustrated in Figure~\ref{fig:llm_pipeline}.

\subsection{eXplainable Artificial Intelligence (XAI) Model}

XAI has recently emerged as a key technique for understanding nonlinear relationships in data. It provides transparency and interpretability to the otherwise black-box nature of AI modeling processes. The XAI framework consists of two main components: the \textbf{prediction model} and the \textbf{interpretation model}. The \textbf{prediction model} is responsible for learning the relationship between input features and outputs. Ensemble models, such as boosting algorithms and random forests, are widely used for this purpose. These models are highly effective, as they capture complex relationships without requiring strict assumptions about the data. The \textbf{interpretation model} analyzes how each input feature affects the outputs of the prediction model. Interpretation techniques are generally divided into two categories: \textit{model-specific} and \textit{model-agnostic}. Model-specific techniques rely on the internal structure of algorithms, offering tailored interpretability but with limited applicability. In contrast, model-agnostic techniques are post-hoc methods that can be applied to any model, making them highly flexible. Among model-agnostic techniques, SHapley Additive exPlanations (SHAP) stands out due to its strong theoretical foundation and versatility~\cite{lundberg2017unified}.

In this study, four prediction models were tested: \textit{Multilayer Perceptron (MLP)}, \textit{Random Forest (RF)}, \textit{Light Gradient Boosting Machine (LightGBM)}, and \textit{eXtreme Gradient Boosting (XGBoost)}. The model with the best performance was selected for interpretation. SHAP was then applied to analyze feature contributions, providing both global and local interpretability. The output was \textbf{political affiliation} (1 = conservative, 0 = progressive), modeled through an additive feature function.

The 19 input features were: (1) gender (1 for male, 0 for female), (2) election type (1 for constituency-based elections, 0 for proportional representation), (3) committee type (1 for transportation, 0 otherwise), (4) number of terms elected, (5) electoral population, (6) number of votes, (7) number of invalid votes, (8) area in square kilometers, (9) number of sponsors, (10) number of conservative sponsors, (11) number of progressive sponsors, (12) percentage of conservative sponsors, (13) percentage of progressive sponsors, (14) average number of terms elected, (15) number of male sponsors, (16) number of female sponsors, (17) percentage of male sponsors, (18) percentage of female sponsors, and (19) approval status.

\subsubsection{Prediction Models}

\paragraph{1) MLP} 
MLP is a feedforward neural network architecture composed of an input layer, one or more hidden layers, and an output layer~\cite{min2023deep, tolstikhin2021mlp}. It learns complex nonlinear relationships via backpropagation through iterative weight updates. While highly flexible, MLPs often require larger datasets and careful regularization techniques to prevent overfitting.

\paragraph{2) RF} 
RF is an ensemble learning method that constructs multiple randomized decision trees and aggregates their outputs~\cite{breiman2001random}. It introduces randomness via bootstrap sampling and feature selection at each split, offering high robustness and strong resistance to overfitting. RF models are easy to implement, handle both categorical and continuous variables well, and provide interpretable internal measures of feature importance.

\paragraph{3) LightGBM} 
LightGBM is a gradient boosting framework that uses tree-based learning algorithms with histogram optimization and leaf-wise splitting~\cite{ke2017lightgbm}. It is well-suited for large-scale and high-dimensional data, providing efficiency in training time and memory. LightGBM also handles missing values natively and supports categorical features without requiring preprocessing. 

\paragraph{4) XGBoost} 
XGBoost is a scalable and efficient gradient boosting implementation that sequentially builds decision trees to correct prior prediction errors~\cite{chen2016xgboost}. It incorporates regularization terms and supports parallel processing, achieving an excellent balance between performance and interpretability~\cite{chikaraishi2020possibility, jin2023container, lee2024explainable, kwak2024impact}. XGBoost is widely used in real-world applications, machine learning competitions, and applied research due to its high predictive accuracy and modeling flexibility.

\subsubsection{Interpretation Model}
SHAP is a widely used model-agnostic technique for interpreting the outputs of black-box machine learning models. Based on cooperative game theory, SHAP values represent the average marginal contribution of each feature across all possible feature combinations, ensuring fairness and consistency in attribution (Lee, 2024). This allows SHAP to provide a clear, mathematically grounded explanation of how individual features influence model predictions. By decomposing the output into additive contributions, SHAP helps users understand the role of each input in the final prediction. The framework operates iteratively, estimating feature effects relative to baseline value (Lundberg and Lee, 2017). For example, it can effectively interpret categorical features such as political affiliation (e.g., 1 for conservative, 0 for progressive) using an additive feature function.

\section{Results}
\subsection{Keyword Extraction}

In the initial keyword extraction experiment, two scenarios were compared: extracting keywords directly from Korean summaries and extracting keywords after translation into English. Validation was conducted through manual review of a randomly selected 10\% sample, all of which were confirmed to be accurate. Direct extraction from Korean yielded 2,769 keywords, many of which were misclassified due to morphological attachments and partial matches. In contrast, after translating the summaries into English, the LLM produced a cleaner set of approximately 2,000 transportation-related keywords. The improved clarity stemmed from English’s simpler noun forms and clearer word boundaries, which helped the model separate genuine transportation terms from incidental markers or suffixes.

The translation stage proved crucial in minimizing keyword extraction errors caused by the agglutinative nature of Korean. In Korean, grammatical particles and auxiliary markers often attach directly to the ends of words, complicating accurate keyword identification. For instance, the Korean word \textit{doro} (road) may be confused with \textit{jedoro} (as a system), where the suffix \textit{-ro} functions as a particle and not part of the transportation term. Such cases lead to false positives when keywords are matched based on character sequences rather than semantic context.

A notable observation was that direct extraction from Korean often misclassified keywords due to morphological ambiguity. Additionally, words with different particle endings were treated as separate keywords, inflating the count and reducing semantic precision. In contrast, English translation prior to extraction yielded more consistent and contextually accurate keywords, as English syntax more clearly separates base nouns from modifiers or particles.

\begin{figure}[htbp]
  \centering
  \vspace{-10pt}
  \includegraphics[width=0.7\linewidth]{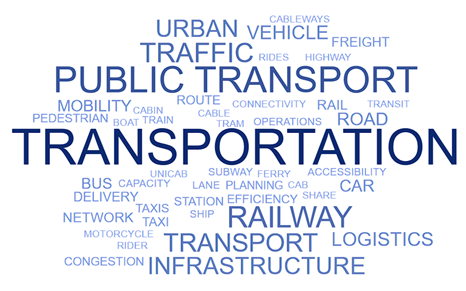}
  \vspace{-0pt}
  \caption{Transportation Word Cloud from Bill data}
  \vspace{-0pt}
  \label{fig:keyword_cloud}
\end{figure}

This improvement is illustrated in Figure~\ref{fig:keyword_cloud}, which presents the extracted keywords from the bill dataset in the form of a word cloud. Larger and darker words indicate higher frequency, reflecting the core vocabulary associated with transportation-related legislation.

\subsection{LLM Classification Outcomes}

After establishing a more accurate keyword set through translation and extraction, a three-stage filtering process was implemented to progressively refine the dataset and isolate bills closely aligned with transportation. Each stage was designed with a specific objective: the first ensured broad coverage, the second introduced contextual rigor, and the third guaranteed thematic relevance.

\textbf{Step 1: Keyword-based Filtering.} The first step involved identifying all bills containing at least one of the extracted keywords. This broad instruction filtered the dataset from 23,655 to 3,874 bills, though many included only incidental or metaphorical references to transportation. While effective for quickly narrowing the scope, this stage exhibited low thematic precision due to its reliance on simple keyword matching.

\textbf{Step 2: Sentence-based Filtering.} In the second step, a more sentence-sensitive criterion was applied. The LLM evaluated whether the identified keywords were used in a substantive transportation context—that is, whether the sentence was not merely containing a keyword, but conveying transportation-related content. As a result, the dataset was reduced to 902 bills. This stage improved relevance by considering the linguistic environment around the keywords, though some marginally related bills remained.

\textbf{Step 3: Context-based Filtering.} The third step filtered for bills where the main content of the summary explicitly focused on transportation policy. Unlike the previous step, which considered sentence-level context, this stage evaluated the overall thematic focus of the bill summary. The LLM was prompted to determine whether transportation was the central policy topic, rather than a secondary or incidental issue. Under this strictest criterion, the dataset was further narrowed to 577 bills, retaining only those clearly centered on transportation policy, infrastructure, or services. This produced a highly precise, transportation-focused subset suitable for downstream analysis.

Overall, the three-stage process reduced the dataset from 23,655 to 577 final records, achieving a 97.6\% reduction. Specifically, 3,874 bills (16.4\%) were selected at the keyword-based stage, 902 bills (3.8\%) at the sentence-based stage, and 577 bills (2.4\%) at the context-based stage. The results of these filtering steps are summarized in Table~\ref{tab:filtering_steps_row}.

\begin{table}[h]
\centering
\caption{Stepwise Filtering Outcomes}
\label{tab:filtering_steps_row}
\begin{tabularx}{\textwidth}{l l X}
\toprule
\textbf{Category} & \textbf{Stage} & \textbf{Description} \\
\midrule
\multirow{3}{*}{Filter Count} 
  & Keyword-based selection & 3,874 \\
  & Sentence-based selection & 902 \\
  & Context-based selection  & 577 \\
\midrule
\multirow{3}{*}{Prompt Example} 
  & Keyword-based selection & “Identify all legislative bills containing at least one transportation-related keyword.” \\
  & Sentence-based selection & “Among the previously selected bills, determine which exhibit meaningful transportation-related context in their summaries.” \\
  & Context-based selection  & “Select only those bills in which transportation constitutes the primary subject or intended legislative outcome.” \\
\midrule
\multirow{3}{*}{Filter Condition} 
  & Keyword-based selection & At least one keyword appears in the summary \\
  & Sentence-based selection & Transportation-related context appears in the summary \\
  & Context-based selection  & The main context of the summary is transportation \\
\midrule
\multirow{3}{*}{Relevance} 
  & Keyword-based selection & Low (lexical matching) \\
  & Sentence-based selection & High (contextual consideration) \\
  & Context-based selection  & Very high (core thematic relevance) \\
\midrule
\multirow{3}{*}{Scope} 
  & Keyword-based selection & Broad \\
  & Sentence-based selection & Medium \\
  & Context-based selection  & Narrow \\
\midrule
\multirow{3}{*}{Error Possibility} 
  & Keyword-based selection & Potential for mislabeling due to redundancy or incidental use of keywords \\
  & Sentence-based selection & Possibility of filtering irrelevant bills \\
  & Context-based selection  & Possibility of excluding broadly related transportation bills \\
\bottomrule
\end{tabularx}
\end{table}

With the data set filtered through the three-stage process, data pre-processing was conducted to examine differences among political parties in terms of lawmakers' participation in transportation-related bills. Since each bill involved multiple lawmakers, the data was structured to include information on participating lawmakers, their party affiliations, and constituency characteristics.

The dataset included three main feature categories: \textit{legislator attributes}, \textit{constituency attributes}, and \textit{bill attributes}. Legislator attributes consisted of political orientation, gender, election method, committee membership, primary or co-sponsor status, and number of terms elected. Constituency-level features included the electoral population, number of votes, number of invalid votes, and total land area. Bill-level attributes captured the total number of sponsors per bill, broken down by party (progressive, conservative) and gender (male, female), as well as the corresponding proportions.

The preprocessed dataset contained a total of 7,872 bill participation instances across 577 transportation-related bills. Of these, 5,030 were from progressive party members and 2,842 from conservative party members. In terms of gender, 6,510 participants were male and 1,362 were female. Regarding election method, 6,735 lawmakers were elected through constituency-based elections, while 1,137 were elected via proportional representation.

For constituency attributes, the average electoral population was 147,555. The average number of votes cast and invalid votes were 97,317 and 1,264, respectively. As for bill-level statistics, each bill had an average of 26.3 sponsors. Of these, progressive sponsors accounted for 16.1 on average, while conservative sponsors averaged 10.1 per bill.

These statistics highlight notable differences in lawmaker participation based on political party, gender, and election method. Progressive lawmakers participated significantly more frequently than conservatives. Male lawmakers accounted for over four times as many participation as female lawmakers. Constituency-based lawmakers were also much more active than their proportional-representation counterparts. These descriptive insights underscore the varying patterns of legislative engagement across political, demographic, and electoral lines and form the basis for further analysis in this study. The descriptive statistics of proposal participation are shown in Table~\ref{tab:proposal_stats}.

\begin{table}[h]
\centering
\caption{Descriptive Statistics of Proposal Participation}
\label{tab:proposal_stats}
\begin{tabular}{llccc}
\toprule
\textbf{} & \textbf{} & \textbf{Count} & \textbf{Average} & \textbf{Data type} \\
\midrule
\multicolumn{2}{l}{Total number of bill participation} & 7,872 & -- & -- \\
\midrule
\multirow{9}{*}{\textbf{Legislator attributes}} 
  & Ideology: Conservative & 2,842 & -- & Categorical \\
  & Ideology: Progressive & 5,030 & -- & Categorical \\
  & Gender: Male & 6,510 & -- & Categorical \\
  & Gender: Female & 1,362 & -- & Categorical \\
  & Election type: Constituency-based & 6,735 & -- & Categorical \\
  & Election type: Proportional & 1,137 & -- & Categorical \\
  & Committee: Transportation & 4,248 & -- & Categorical \\
  & Committee: Others & 3,624 & -- & Categorical \\
  & Sponsor status & 579 & -- & Categorical \\
  & Co-sponsor status & 7,293 & -- & Categorical \\
  & No. of terms elected & -- & 2.25 & Numeric \\
\midrule
\multirow{4}{*}{\textbf{Constituency attributes}} 
  & Electoral population & -- & 147,555 & Numeric \\
  & Number of votes & -- & 97,317 & Numeric \\
  & Number of invalid votes & -- & 1,264 & Numeric \\
  & Area (km\textsuperscript{2}) & -- & 322 & Numeric \\
\midrule
\multirow{11}{*}{\textbf{Bill attributes}} 
  & No. of sponsors & -- & 26.3 & Numeric \\
  & No. of conservative sponsors & -- & 10.1 & Numeric \\
  & No. of progressive sponsors & -- & 16.1 & Numeric \\
  & \% of conservative sponsors & -- & 0.39 & Numeric \\
  & \% of progressive sponsors & -- & 0.61 & Numeric \\
  & Avg. No. of terms elected & -- & 2.2 & Numeric \\
  & No. of male sponsors & -- & 21.53 & Numeric \\
  & No. of female sponsors & -- & 4.76 & Numeric \\
  & \% of male sponsors & -- & 0.826 & Numeric \\
  & \% of female sponsors & -- & 0.174 & Numeric \\
  & Approval: Accept & 916 & -- & Numeric \\
  & Approval: Reject & 6,956 & -- & Numeric \\
\bottomrule
\end{tabular}
\end{table}

\subsection{Prediction Model Performance}

With the selected set of transportation-related bills identified through the LLM-based classification process, a suite of explainable AI (XAI) models was developed to explore the relationships between input features and political party affiliation. As described earlier, the XAI framework consists of two components: a prediction model that learns the mapping between inputs and outputs, and an interpretation model that explains these relationships based on the prediction outcomes.

In this study, four prediction models were implemented, i.e., MLP, RF, LightGBM, and XGBoost, to capture the associations between political party affiliation (conservative vs. progressive) and a range of input attributes, including legislator characteristics, constituency demographics, and bill-level indicators.

All models were trained and evaluated using a consistent experimental protocol to ensure fair comparison. The dataset was randomly divided into training and testing sets with an 85:15 ratio, using a fixed random seed of 42 to ensure reproducibility. Hyperparameter optimization was performed via grid search with five-fold cross-validation. For each iteration, four folds were used for training and the remaining fold for validation. This process was repeated five times, and the average performance was used to determine the optimal configuration for each model, thereby minimizing variance and reducing the risk of overfitting.

The final hyperparameter settings for each model were as follows:
\begin{itemize}
  \item \textbf{MLP}: hidden layer size = 16; activation function = ReLU; solver = Adam
  \item \textbf{Random Forest}: number of estimators = 500; maximum depth = 8; criterion = Gini impurity
  \item \textbf{LightGBM}: number of iterations = 500; learning rate = 0.12; number of leaves = 16
  \item \textbf{XGBoost}: number of iterations = 500; learning rate = 0.15; maximum depth = 16
\end{itemize}

Model performance was evaluated using three standard classification metrics: precision, recall, and F1 score. Precision quantifies the proportion of correctly predicted positive observations among all positive predictions. Recall measures the proportion of actual positive cases that were correctly identified. The F1 score provides a harmonic mean between precision and recall, offering a balanced evaluation when the two metrics diverge. The mathematical formulas are given in Equations~(3)--(5).

\begin{align}
\text{Precision} &= \frac{TP}{TP + FP} \tag{3} \\
\text{Recall} &= \frac{TP}{TP + FN} \tag{4} \\
F_1 \ \text{score} &= 2 \times \frac{\text{Precision} \times \text{Recall}}{\text{Precision} + \text{Recall}} \tag{5}
\end{align}

\noindent
where $TP$ is the true positive, $FP$ is the false positive, $TN$ is the true negative, and $FN$ is the false negative.

The models were trained on 85\% of the dataset and validated on the remaining 15\%, with both subsets randomly sampled. The training set consisted of 8,048 records out of a total of 10,060, while the test set comprised 2,012 records. Each model was evaluated over ten independent runs using different random seeds. The reported metrics represent the average performance across these ten runs, with 95\% confidence intervals (CIs) included to demonstrate the stability and robustness of each model.

The MLP model achieved scores of 0.877 for precision, 0.853 for recall, and 0.865 for F1 score. The Random Forest (RF) model showed improved results, with values of 0.932 for precision, 0.916 for recall, and 0.924 for F1 score. The LightGBM model further enhanced these metrics, achieving 0.961 precision, 0.948 recall, and 0.955 F1 score. Finally, the XGBoost model outperformed all others, with the highest scores of 0.977 for precision, 0.979 for recall, and 0.978 for F1 score.

These results indicate that all models demonstrated strong predictive performance and are suitable for use in interpretation tasks. However, XGBoost consistently achieved the best results across all evaluation metrics and was thus selected as the final model for interpreting the relationships between political party affiliation and input attributes. The comparative performances of the four models are summarized in Table~\ref{tab:prediction_performance}.

\begin{table}[h]
\centering
\caption{Performance of the prediction model}
\label{tab:prediction_performance}
\begin{tabular}{lccc}
\toprule
\textbf{} & \textbf{Precision (95\% CI)} & \textbf{Recall (95\% CI)} & \textbf{F1 Score (95\% CI)} \\
\midrule
MLP & 0.877 $\pm$ 0.005 & 0.853 $\pm$ 0.006 & 0.865 $\pm$ 0.005 \\
RF & 0.932 $\pm$ 0.004 & 0.916 $\pm$ 0.004 & 0.924 $\pm$ 0.004 \\
LightGBM & 0.961 $\pm$ 0.003 & 0.948 $\pm$ 0.004 & 0.955 $\pm$ 0.003 \\
\textbf{XGBoost} & \textbf{0.977 $\pm$ 0.002} & \textbf{0.979 $\pm$ 0.003} & \textbf{0.978 $\pm$ 0.002} \\
\bottomrule
\end{tabular}
\end{table}

To further evaluate the classification performance of the XGBoost model, a confusion matrix was constructed to assess its ability to distinguish between conservative and progressive party members, as shown in Table~\ref{tab:confusion_matrix}. The model correctly classified 986 conservative and 982 progressive legislators. However, 23 progressive members were misclassified as conservative, while 21 conservative members were misclassified as progressive.

Based on the confusion matrix values, precision, recall, and F1 score were calculated. For the conservative class, precision was computed as $\frac{986}{986 + 23} = 0.977$, and recall as $\frac{986}{986 + 21} = 0.979$. The F1 score, representing the harmonic mean of precision and recall, was derived as $\frac{0.977 \times 0.979}{0.977 + 0.979} = 0.978$.

These results indicate that the XGBoost model demonstrated excellent classification performance, achieving consistently high precision, recall, and F1 scores for both ideological groups.

\begin{table}[h]
\centering
\caption{Confusion matrix of XGBoost model}
\label{tab:confusion_matrix}
\begin{tabular}{lcc}
\toprule
\textbf{} & \textbf{Actual conservative party (0)} & \textbf{Actual progressive party (1)} \\
\midrule
Predicted conservative party (0) & 986 & 23 \\
Predicted progressive party (1)  & 21  & 982 \\
\bottomrule
\end{tabular}
\end{table}

\subsection{XAI Interpretations}
\subsubsection{Feature Importance}

\begin{figure}[htbp]
  \centering
  \includegraphics[width=1.0\linewidth]{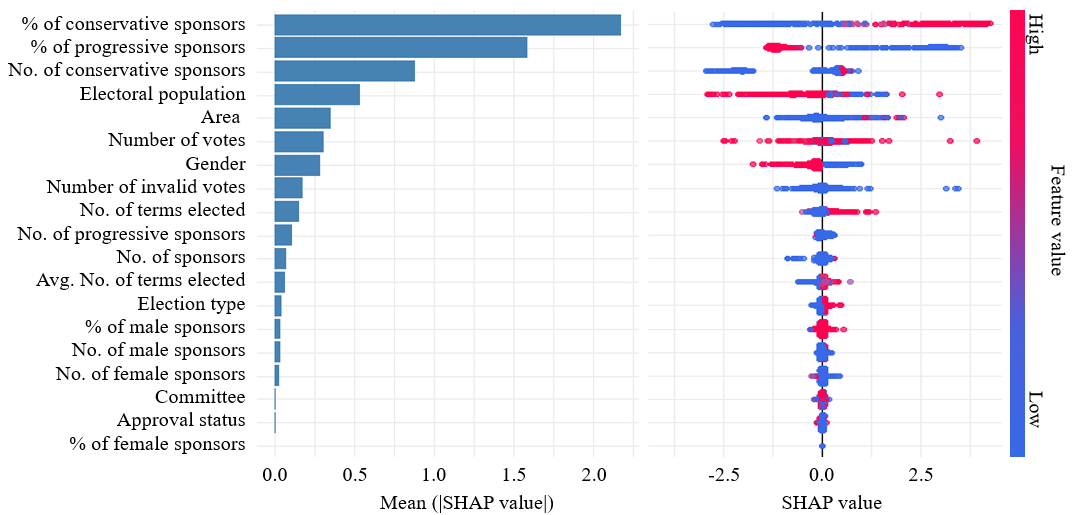}
  \caption{Results of feature analysis.}
  \label{fig:shap_summary}
\end{figure}

Figure~\ref{fig:shap_summary} presents the SHAP values of the 19 features used in the XGBoost model. The features are ranked according to their mean absolute SHAP values, which represent their average contribution to the model's predictions. The results show that the most influential feature was the percentage of conservative sponsors, followed by the percentage of progressive sponsors and the number of conservative sponsors. These variables play a critical role in determining political affiliation. Constituency attributes such as electoral population and area also exhibited high importance, suggesting that district-level characteristics substantially influence classification outcomes. Moderate contributions were observed from the number of votes and legislator gender, reflecting the relevance of voting behavior and demographic profiles. Less influential—but still informative—features included the number of invalid votes, the number of terms served, and the average number of terms served. The SHAP summary plot revealed clear patterns in feature effects: higher values of conservative or progressive sponsorship were positively associated with their respective party classifications. Conversely, variables such as the number and percentage of female sponsors showed lower SHAP values, indicating limited influence on the model's output.

Overall, these results suggest that the key drivers of political classification in the XGBoost model include sponsor composition by party, electoral population, and district area. Specifically, conservative legislators are more likely to sponsor or co-sponsor bills introduced by other conservatives, while progressive legislators tend to support legislation from their own party. This indicates a strong ideological alignment in legislative activities.

Additionally, conservative legislators were more commonly associated with constituencies characterized by smaller populations and larger geographical areas, while progressive legislators were linked to denser, urban districts. These findings align with broader political patterns: conservative parties generally perform better in rural or low-density areas, whereas progressive parties gain stronger support in urban and high-density regions~\cite{klein2022political, legacy2019consensus, mcandrews2015politics, sciara2024state}. For instance, conservatives often advocate for highway expansion and traditional infrastructure to serve rural mobility needs, whereas progressives prioritize sustainable urban transit and multimodal infrastructure.

\begin{wrapfigure}{r}{0.55\textwidth}
  \centering
  \vspace{-0pt}
  \includegraphics[width=0.53\textwidth]{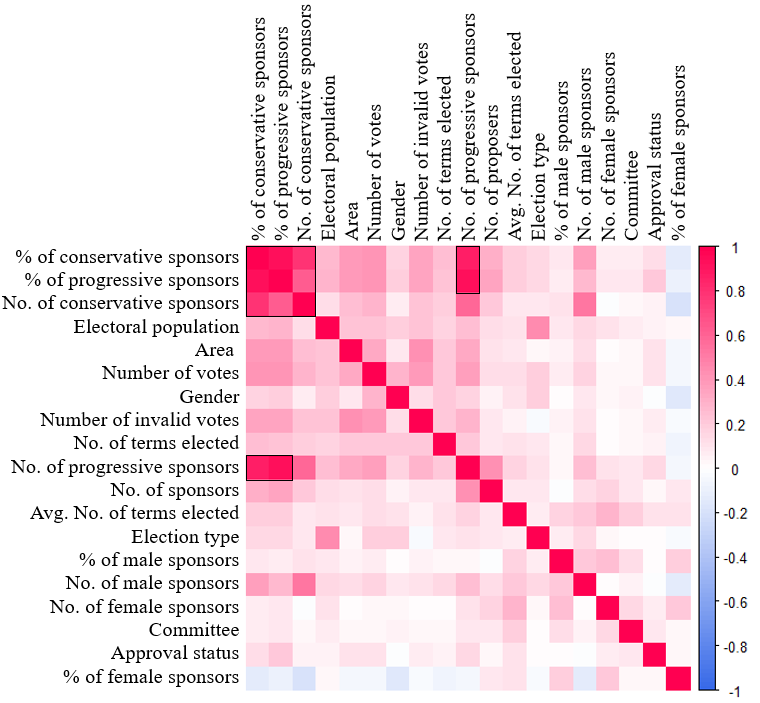}
  \vspace{-10pt}
  \caption{Relationship between SHAP values.}
    \vspace{-15pt}
  \label{fig:shap_corr}
\end{wrapfigure}

One of the key advantages of XAI is its ability to uncover and interpret relationships among input features. To explore these interdependencies, it is essential to identify a subset of features that significantly influence model outputs. As part of analysis, a correlation was estimated focusing on the top four features—out of the 19 total—that exhibited the highest impact on classification outcomes.

Figure~\ref{fig:shap_corr} visualizes the relationships among all 19 features based on their SHAP values. The results reveal that the top four features—percentage of conservative sponsors, percentage of progressive sponsors, number of conservative sponsors, and number of progressive sponsors—were strongly positively correlated, each exhibiting Pearson correlation coefficients greater than 0.6. These findings suggest that sponsor-related features are not only individually important, but also exhibit interrelated patterns that collectively influence the model’s prediction of political affiliation.

\subsubsection{Feature Dependency Analysis}

\begin{figure}[htbp]
  \centering
  \includegraphics[width=\linewidth]{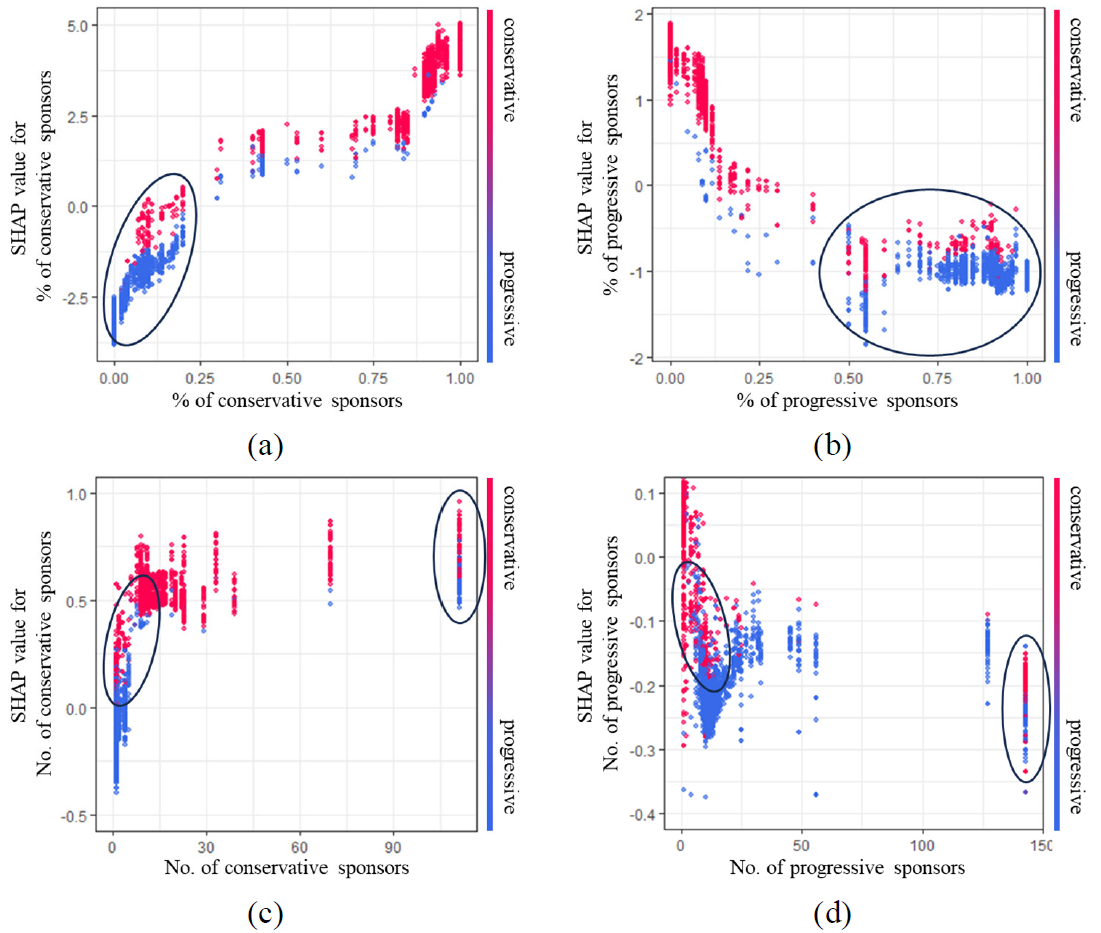}
  \caption{Result of SHAP dependency analysis: (a) the proportion of conservative sponsors, (b) the proportion of conservative sponsors, (c) the number of conservative sponsors, (d) the number of progressive sponsors.}
  \label{fig:shap_depend}
\end{figure}

The feature dependency results in Figure~\ref{fig:shap_depend} show how key variables influence predictions of party affiliation. The SHAP plots illustrate both marginal effects and interactions among conservative and progressive sponsor patterns.

In Figure~\ref{fig:shap_depend}(a) and (b), the proportion of conservative sponsors increases SHAP values, indicating stronger conservative predictions—especially beyond 50\%. In contrast, a higher proportion of progressive sponsors leads to negative SHAP values, signaling progressive alignment.

Figure~\ref{fig:shap_depend}(c) and (d) show that conservative sponsor counts above 90 consistently yield positive SHAP values, while lower progressive counts (under 50) are associated with negative SHAP values. At high progressive counts (150+), SHAP values stabilize near zero.

These results suggest that both the proportion and volume of sponsorship influence ideological alignment. Importantly, instances where conservative legislators appear in progressive-dominated contexts indicate potential bipartisan cooperation.

\section{Discussion}

This study aimed to understand the impact of political ideologies on transportation-related legislative bills by developing an innovative framework that combines LLMs and XAI techniques. Specifically, GPT-4 was used to efficiently process and classify legislative texts, ensuring contextual relevance, while XGBoost was employed to model the relationships between political party affiliations and transportation-related bill proposals. SHAP analysis was further used to interpret the influence of input features on model predictions.

This study makes several contributions to both research and practice. 

First, it developed a structured framework for transportation policy analysis using LLMs. The stepwise classification approach filtered bills based on keywords, semantic content, and contextual relevance. The results showed that LLMs enabled significantly more precise classification, identifying 577 transportation-related bills, compared to 3,874 identified using keyword-only methods---an improvement of approximately 85\%. Leveraging the deep contextual understanding of GPT-4, this study introduced a scalable approach to mining large-scale legislative data. This aligns with recent work highlighting the strengths of LLMs in extracting policy-relevant information from complex text corpora~\cite{tornberg2024large, wan2024}.

Second, the study employed AI-based modeling techniques to analyze relationships between party affiliation and various attributes related to legislators, districts, and bill sponsorship. Among the models tested, XGBoost showed the highest accuracy in classification, consistent with findings from prior studies on tabular data classification~\cite{hak2021estimating, ramraj2016experimenting}. These results suggest that tree-based ensemble methods are particularly effective in modeling political behavior and legislative dynamics.

Third, the study applied SHAP values to interpret model predictions. The results indicated that the proportion and number of progressive and conservative sponsors were the most influential features in predicting party alignment. This confirms that legislative composition often mirrors broader ideological preferences. For instance, Klein et al.~\cite{klein2022political} and Nixon and Agrawal~\cite{nixon2019would} found that political ideology significantly influences public support for transportation policies. Democrats tend to support urban transit and congestion pricing, while Republicans are more likely to favor highway investment and oppose fuel taxes. This study extends those findings to the legislative context, suggesting that similar ideological divides shape bill sponsorship patterns and policy formation. Moreover, district-level attributes such as electoral population and land area were also influential, indicating that geography plays a role in shaping legislative behavior.

Finally, the results offer practical implications for transportation policymakers. The analysis showed that South Korea’s 21st National Assembly exhibited both bipartisan cooperation and ideological division. While conservative and progressive parties often collaborated on broadly supported policies, they tended to avoid participation in each other’s proposals on ideologically contentious issues. This pattern reflects broader policy diffusion processes in which decision-makers observe the behavior of ideological peers~\cite{chae2008conservatives, hacker2019policy, trapenberg2015politics}. Understanding such dynamics may help design transportation policies that are more likely to gain cross-party support, as recommended by recent work on policy alignment and collaborative governance~\cite{tiznado2025towards}.

\section{Conclusion}
\paragraph{Conclusion.}  
We proposed \textbf{LegiGPT}, a novel pipeline that combines LLMs and XAI techniques to analyze the influence of political ideologies on transportation-related legislation. Using GPT-4, the framework filters legislative texts with high contextual accuracy, followed by classification using XGBoost and interpretation with SHAP. Compared to traditional keyword matching, LegiGPT improves thematic precision by over 85\%. SHAP results highlight that party-affiliated sponsorship patterns and constituency characteristics (e.g., electoral population, area size) are strong predictors of political alignment in transportation policy. This integrative pipeline provides a scalable, interpretable method for legislative analysis.

\paragraph{Limitations.}  
LegiGPT is limited in several respects. First, the analysis is restricted to South Korea’s 21st National Assembly, which may reduce generalizability to other political systems. Second, the dataset reflects static legislative behavior and does not capture temporal dynamics such as public opinion shifts or party realignment. Third, the framework may require prompt adaptation and validation when applied to other languages or policy domains. Finally, the reliance on pre-trained LLMs introduces risks of semantic bias, especially in politically sensitive contexts.

\newpage
\bibliographystyle{unsrt}
\bibliography{references}

\end{document}